# Robot joint characterisation and control using a magneto-optical rotary encoder

Yunlong Guo[1], John Canning[2, 3] *, Zenon Chaczko[4], Gang-Ding Peng[1]

[1] *School of Electrical Engineering and Telecommunications, Engineering Faculty, University of New South Wales, NSW, 2052, Australia*

[2] *Laseire Consulting Pty Ltd, Sydney, NSW, Australia*

[3] *Department of Mathematics and Physics, University of South China, China*

[3]*DIVE IN AI, Wrocław, Poland*

* *canning.john@outlook.com*

*Abstract*—**A robust and compact magneto-optical rotary encoder for the characterisation of robotic rotary joints is demonstrated. The system employs magnetic field-induced optical attenuation in a double-pass configuration using rotating nonuniform magnets around an optical circulator operating in reflection. The encoder tracks continuous 360° rotation with rotation sweep rates from $v$ = 135 °/s to $v$ = 370 °/s, and an angular resolution of $\Delta\theta$ = 0.3°. This offers a low-cost and reliable alternative to conventional robot rotation encoders while maintaining competitive performance.**

## I. INTRODUCTION

ROTARY encoders convert rotation into electromagnetic signals, most commonly electrical. Examples include precision monitoring and control of steering wheels [1], [2], motors of autopilot vehicles [2], [3], robotics [4], [5], and prosthetic arms [6]. In robotics, the encoder is a crucial part of the positional feedback needed to perform precision movements. Despite widespread use, electrical rotary encoders suffer from a number of challenges that can limit applications [7].

Electronic magnetic encoders are cheaper, smaller, and immune to contaminants, but often suffer from external electromagnetic field interference [8]. Field interference is typically removed with an appropriate differential design that undoes the influence of external interferences and is tailored for each application. The simplest configurations employ potentiometers, where angular displacement modulates electrical resistance, producing corresponding changes in voltage [9]. The capacitor-based rotation encoder is commonly used in steering wheels with disk-shaped transmitting and receiving electrodes mounted on a shaft [10], [11], [12], [13]. Another popular approach is the inductive rotation encoder, which employs self-inductance of the coil or mutual inductances of multiple coils [14], [15], [16], [17] to read rotation angle differences. However, those encoders require sophisticated rotary design, and the quality of coil winding can significantly influence its performance. Applications involving human-like robots and specialised robots—particularly surgical robots—demand reliable and accurate rotation sensing data for encoding during operation [18], [19]. In most cases, environmental variations can be effectively mitigated through differential cancellation between two or more sensors, which is critical for useful encoders. The data transmission of these devices involves electrical cabling, where some of the largest interference and shielding challenges can lie.

Optical-based encoders, in principle, offer higher precision, are themselves mostly free from electromagnetic interference (or e.m.i), and are stable over long-term use. Common examples of bulk optical encoders employ a disc with different optical patterns or magnetic field properties along its edge, an optical transmitter, and an optical receiver [20], [21], [22], [23]. The rotation angular information is read by the reflected information disc with different patterns along its rotation edge. Unfortunately, this technology is sensitive to environmental contamination, such as dust between the disc and the optical receiver, increasing angular detection error. It also requires differential sensing to reduce the impact of mechanical variations such as directional vibrations. In addition, the majority of these still transmit data electronically via electrical cable, with all the distance and energy limitations within and outside of a robotic body this brings. Overall, these do not offer many advantages over lower-cost electronic devices, explaining why their integration as encoders in robotic and other systems has not been as widespread.

A significant disadvantage of these electronic and optical-based devices is the need for electrical cabling to deliver and receive power and control and data signals – this cabling is exposed to external e.m. interference and may require shielding in many applications, particularly where dense, moving cabling is present, such as in advanced network robot nervous systems.

Optical fibre waveguide-based encoders potentially offer a solution to addressing these issues, including replacing the electrical cables with optical fibres. These will be free of electromagnetic field interference and can be operated remotely at much larger distances than electronic cabling. They can connect encoders and indeed other sensor devices directly in various distributed formats along the fibre without any requirement for electrical cabling. However, fibre-based devices alone have limits, including cost. Specialty fibres such as structured, Fresnel, or PCFs (photonic crystal fibre) [24], [25], [26], [27], helical [28], [29], and multicore fibres[30] have been used for rotation sensing, where light transmits into different cores with twisting. These sensors are limited as encoders. Whilst they can provide higher angular resolution, repeated mechanical twisting will limit the fibre's operational range and the physical range that can be measured. Typically, these sensors are confined below rotation angles of $\theta < 180°$ to mitigate against breaking. More generally, the use of specialty fibres instead of standard telecommunications fibres makes these sensors expensive, discouraging efforts to translate them into rotary encoders.

In this work, an approach that overcomes these obstacles by combining the benefits of a compact nonlinear, micro-optical arrangement with standard telecom fibre is proposed, modifying an existing telecommunications component that is well established, having enabled the internet: the optical circulator. The circulator relies on the inherent nonlinear Faraday rotation of light coupled with adjustment of polarisation eigenstates using micro-prisms all within a standard, low-cost telecommunications component having input and output single-mode standard telecommunications optical fibres. Through modification, the nominally insensitive device can be made sensitive to external magnetic fields. When interrogated optically, the introduced circulator response to these fields is encoded into the optical domain and carried by optical fibre to send to a photodetector and receiver. Figure 1 (c) summarises the proposed design. Important for this work, Kovar is an excellent packaging material which reduces variation in physical changes in property with temperature, a reason why it has been used in many device packages of scientific consequence [31], [32]. For normal operation where the Faraday rotator response in a circulator is saturated, a circulator is immune to magnetic fields. However, Kovar is ferromagnetic, and, because of potential induction, it is not used where magnetic fields may be present and any properties are altered. In this work, we will show how this can be taken advantage of. A double pass configuration is used where the transmitted input light is reflected into the optical circulator (Figure 1(a) and (b)) using an in-fibre reflector. This may be a fibre Bragg grating for wavelength division multiplexing of encoders or, for the work presented here, a broadband Au reflecting film deposited onto the end face of an optical fibre. The device is then placed within a non-uniform rotary magnetic field that moves around it. From previous work, the magnitude of this magnetic field is selected to reverse the pre-existing saturation of the circulator so that it can be made sensitive to external fields [33], [34]. This sensitivity generates attenuation of reflected light, which passes through the circulator a second time to further enhance the response as the magnet rotates around the circulator. By appropriate design of the magnetic field configuration over the circulator, full $\theta = 360°$ rotation with high angular resolution is obtained.

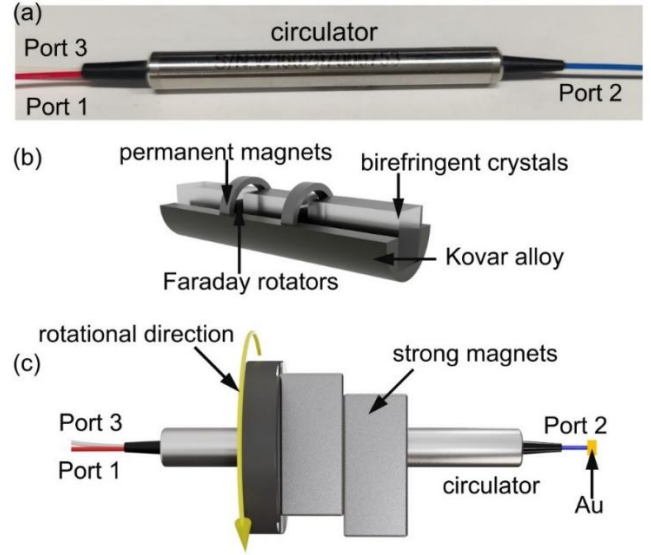

**Fig. 1.** The circulator-based optical rotation encoder: (a) a commercially available circulator; (b) the interior has a Kovar alloy mount supporting the yttrium iron garnet (YIG) Faraday rotators, the birefringent displacement prisms that align rotated light with input and output fibres, and pre-existing magnets that saturate the nonlinear YIG response to make the circulator insensitive to external fields as required for a telecom's transmission router. These elements are coated with thin films to reduce spurious reflections; (c) the circulator-based rotation encoder is composed of an optical circulator and two additional magnets, ($B_{add}$), to pull back the saturation field and make the circulator respond to external magnetic fields. A reflector at Port 2 returns the input signal from Port 1, which is transmitted to Port 3. Rotation of the magnets is illustrated with a yellow arrow.

## II. EXPERIMENT AND RESULT

Additional strong permanent magnets, shown in Fig. 1(c), placed along the circulator were used to undo the magnetic field generated by the permanent magnets inside the circulator shown in Fig. 1 (b). This pulls back the saturated nonlinear response of the YIG crystals to make the circulators sensitive to external fields, seen as changes in optical attenuation of transmitted light. Optimal sensitivity to additional external changes in the magnetic field becomes possible. This was previously used to demonstrate a novel magnetic field sensor [33] and, when combined with an electromagnet, an electrically tuneable attenuator [34], offering an alternative to thermal or mechanical-based attenuators currently used in telecommunications. Considering only the case for an aligned field, the single pass attenuation, $\Delta\alpha$, along the fibre axis reduces to:

$$\Delta\alpha = A\Delta\theta = 2AV_{eff}l_c(\vec{B}_s - \vec{B}_{add}) \quad (1)$$

where $A$ is the scaling constant between attenuation and rotation angle, $V_{eff}$ is the Verdet constant of the Faraday rotator crystals, a material property that quantifies the magneto-optic response, $\vec{B}_s$ is the pre-existing magnetic field within the circulator, and $\vec{B}_{add}$ is the additional magnetic field provided by the strong magnet. In principle magnetic field orientation information is possible to extract for potential application as a gyrometer. In this one-dimensional analysis where the field remains aligned with the fibre axis as it rotates around the circulator, the use of a reflector creates a double pass configuration configurations doubling the interaction along the fibre axis of eq (1):

$$\Delta\alpha = A\Delta\theta = 4AV_{eff}l_c(\vec{B}_s - \vec{B}_{add}) \quad (2)$$

Commercial circulators are manually fabricated, with a half-cylinder Kovar casing designed to facilitate low-cost manual placement of optical elements (Fig. 1(b)). As a result, the circulator is only partially magnetised, shielding certain parts of the device depending on the direction of the applied field $\vec{B}_{add}$. The alignment is neither perfect nor uniform across all products but is manually calibrated as a black box by measuring and reducing device loss to make up for this. This non-uniformity can affect the actual vector field distribution experienced by the Faraday rotators, depending on how $\vec{B}_{add}$ is applied, and so individual device characterisation is warranted. However, the 1D approximation remains experimentally valid so long as the rotation is orthogonal to the circulator and keeps the field at the circulator contents aligned with the propagation axis. The fortunate device rotational asymmetry enables straightforward discrimination of rotation when $\vec{B}_{add}$ is rotated around the circulator, and the generalised attenuation for a single applied $\vec{B}_{add}$ can be reduced to:

$$\Delta\alpha = A\Delta\theta = 4AV_{eff}l_c(\vec{B}_s - (\vec{B}_{add} - \sigma\vec{B}_b)) \quad (3)$$

where $\vec{B}_b$ is the effective magnetic field on Faraday rotators that is distorted by a half-cylinder ferromagnetic Kovar and the existing magnets, and $\sigma$ is the ratio related to the relative position between the applied strong magnets and Faraday rotators, and the magnetic field shielded by the Kovar alloy. The value of the $\sigma$ would vary with the rotation angle of the applied magnets and their position relative to the existing magnets and the half-cylinder Kovar, all of which affect the field profile over the YIG crystals. Practically, this value is obtained by measuring the actual device response.

*A. Circulator magnetic field characterization*

For initial characterisation of the existing magnetic field, $\vec{B}$, within the circulator before any additional magnets are added, $\vec{B}$, is mapped axially along the circulator using a magnetometer on two sides (one side is shielded by the Kovar, the other side is not – see Fig. 1 (b)). Fig. 2 (a) and (b) show these measured profiles demonstrating the role the ferromagnetic Kovar plays in creating an asymmetry that can be exploited. Fig. 2(a) shows the magnetic field on the side where the existing magnetic field is not shielded by the Kovar.

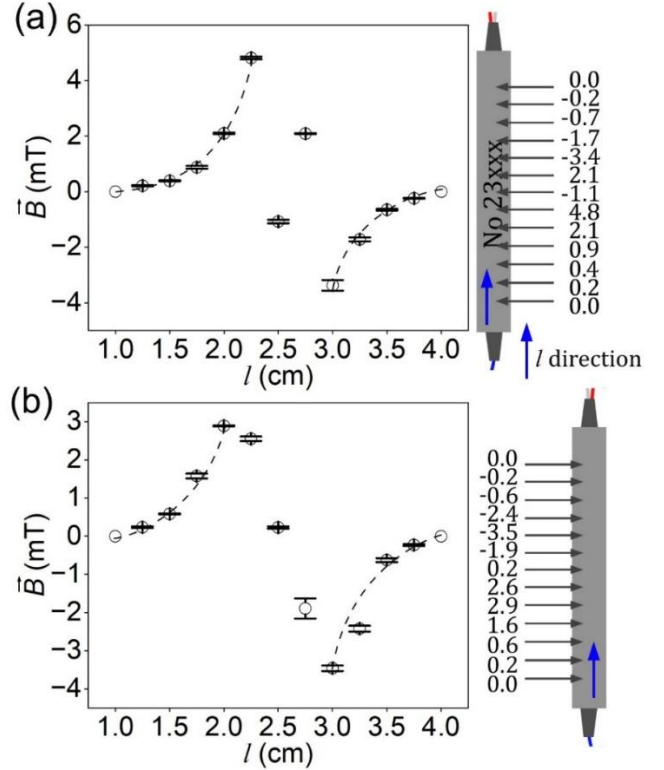

**Fig. 2.** The mapped magnetic field distribution along the circulator surface using a magnetometer placed directly on the surface of the circulator, approximately $d_m \sim 2.3$ mm from the Faraday rotator centres: (a) the magnetic field, $\vec{B}$, on the side unshielded by the Kovar; (b) the magnetic field, $\vec{B}$, distribution on the side shielded by the Kovar. In both cases, the relative position, $l$, along the circulator is measured from the circulator end closest to Port 2.

A peak value $\vec{B} = (4.8 \pm 0.1)$ mT is measured at $l \sim 2$ cm, coinciding with one internal magnet located inside the circulator. Another peak value $\vec{B} = -(3.4 \pm 0.2)$ mT, but opposite in sign is measured at $l \sim 3$ cm, coinciding with the second magnet. The absolute values differ in part from different magnet field strengths, most likely due to human packaging variation in the supply chain that may lead to small differences in, for example, magnet orientation. On the other side of the circulator, these values are reduced to $\vec{B} = (2.9 \pm 0.1)$ mT and $\vec{B} = -(1.9 \pm 0.3)$ mT because the Kovar is an effective shield, in this case nearly 50%. Any positional variation, $\Delta l \sim 0.2$ cm, arises from the resolution error of the field measurements as the magnetometer is translated. These measurements readily identify the location of the YIG crystal within these magnets.

*B. Sensor design and optimization*

Following characterisation of the circulator's intrinsic magnetic field distribution and the identification of the magnets and rotators along the circulator using this novel mapping technique, the rotation encoder system was configured as shown in Fig. 3. An ASE broadband source feeds light through a fibre into the circulator, with the encoder's output detected by a power meter connected to an oscilloscope. Within the rotation encoder assembly, the strong magnets are placed with a magnetic field opposite to the magnetic field generated by the embedded magnets, shown in the inset of Fig. 3. This cancels the magnetic field, reversing the saturated state of the Faraday rotators [33]. This, in turn, makes the circulator much more responsive to external magnetic field variations, leading to changes in attenuation.

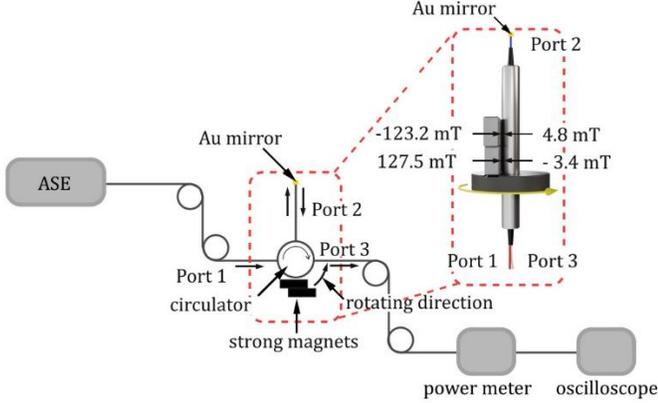

**Fig. 3.** Circulator-based rotation encoder system. Magnets move around the circulator using a DC rotation stage adjusted by varying the applied voltage. These magnets provide the additional external magnetic field $\vec{B}_{add}$ (Negative values represent the reverse magnetic field direction along the optical propagation axis). An amplified spontaneous emission (ASE) source provides light at $l \sim 1550$ nm remotely from the circulator via optical fibre 3 m in length, which is coupled into Port 1. In principle, this measurement can be undertaken via the internet from almost anywhere. The light is coupled into Port 2, where it is then reflected into the circulator to exit from Port 3. The optical transmission attenuation, $a$, is measured from Port 3 on an oscilloscope using a germanium Photodetector, also at a distance $L \sim 3$ m away.

To maximise the attenuation dynamic range and reduce noise, two strong magnets are positioned on the same side of the circulator shown in Fig. 4(a). They are oriented such that their magnetic fields oppose those of the embedded magnets within the device, a configuration that will also reduce other influences because any interference will affect both sets of magnets equally. This is a differential approach to making devices immune to environmental contributions.

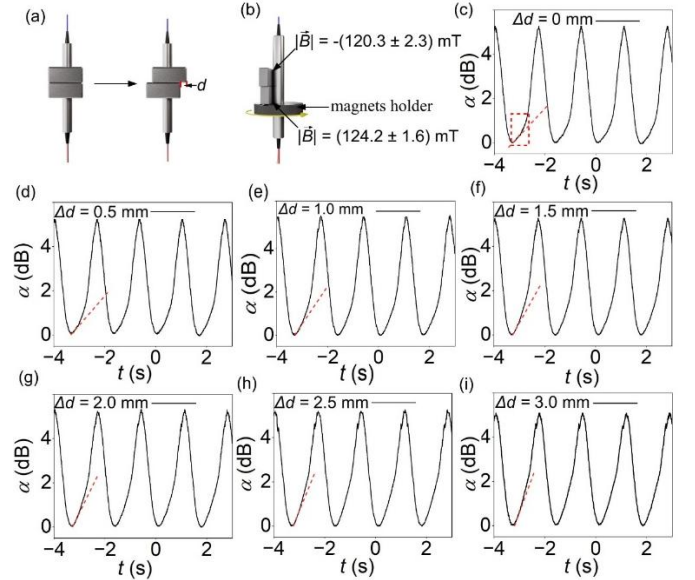

**Fig. 4.** Attenuation, $a$, measured from Port 3 when the added magnets are rotated around the circulator: (a) shows the relative orthogonal position of the magnets from left to right; (b) shows the magnetic field of two magnets (Negative values represent the reverse magnetic field direction) and their relative position with the circulator; (c-i) shows the measured attenuation, $a$, as a function of time, $t$, from Port 3 on an oscilloscope with rotating magnets in different relative orthogonal positions, $d$, moving around the circulator. The temporal resolution was $\Delta t = 5$ ms.

When the added magnets are aligned along the circulator with each other, $\Delta d = 0$ mm, shown in Fig. 4 (a), the measured optical attenuation, $a$, as a function of time, $t$, is shown in Fig. 4 (c - i). It exhibits a low rotational signal per unit angle, $q$, making it difficult to achieve high rotary angle resolution, $\Delta\theta$. When the relative position ($\Delta d$) between the two strong magnets was adjusted in incremental steps of 0.5 mm, an optimum position for rotary angle resolution can be found. When the separation reached $\Delta d = 3.0$ mm (Fig. 3 (a) right side), the rotation encoder achieved an average resolution of $\Delta\theta = 0.3°$.

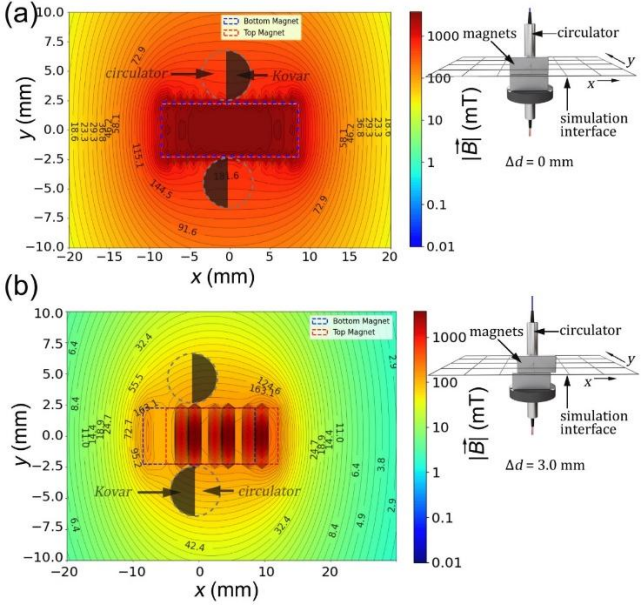

**Fig. 5.** Simulated magnetic field, $\vec{B}$, with different relative magnet displacements with respect to each other: (a) $\vec{B}$ distribution for the aligned magnets where $\Delta d = 0$ mm relative to the circulator and internal Kovar. The diagram represents the relative positions of the circulator before ($\theta = 0°$) and after a 180° rotation ($\theta = 180°$); (b) $\vec{B}$ distribution for the misaligned magnet where $\Delta d = 3.0$ mm.

In order to better appreciate the reasons behind these improvements with small displacements, magnetic field distributions were simulated using Python with open-source packages NumPy [35], [36] and Matplotlib [37]. The simulation parameters of the magnet were as follows: magnet-type N52 Nd with residual magnetic flux density, $|\vec{B}| = |\vec{B}|_r = 1.0$ T, and physical dimensions (17 × 9 × 4.5) mm. The simulated $\vec{B}$ generated by the external magnets (Fig. 4 (a)) is shown in Fig. 5. When $\Delta d = 0$ mm, $\vec{B}$ is relatively uniform, evenly distributed on different sides (left and right sides in Fig. 5 (a) of the circulator). However, when $\Delta d = 3.0$ mm, $\vec{B}$ differs and is asymmetric on the different sides (also left and right sides in Fig. 5 (b)) of the circulator. This variation, especially in the orientation and magnitude of the magnetic vectors that interact with the Faraday rotators, leads to distinct differences in polarisation-induced attenuation variation when rotating. Beyond generating a spatially non-uniform magnetic field, the dual-magnet configuration appears to suppress environmental magnetic noise, which can be explained by balancing the opposite orientation of the internal and external fields appropriately to affect a degree of differential offset, as intended. As shown in Fig. 5, $|\vec{B}| > 20$ mT at distances exceeding the circulator physical diameter, $f = 4.5$ mm. In addition to the differential arrangement, the field magnitude is more than two orders of magnitude stronger than background magnetic noise, whether from Earth's geomagnetic field ($|\vec{B}| \sim 50$ μT) or from typical ambient interference ($|\vec{B}| \sim 1 - 100$ μT).

*C. Stability test and application*

The performance of the rotation encoder was characterised at different rotational velocities, measured here as angle swept per second ($v = \theta/s$), using the encoder system in Fig. 3. A DC rotation stage is used to rotate the magnets around the circulator. Shown in Fig. 6 (b to i), the optical attenuation remains consistent and stable with increasing $v$. The attenuation, $a$, as a function of rotation angle, shown in Fig. 6 (j), demonstrates that errors are confined within a reasonable tolerance: $\Delta a = \pm 0.05$ dB.

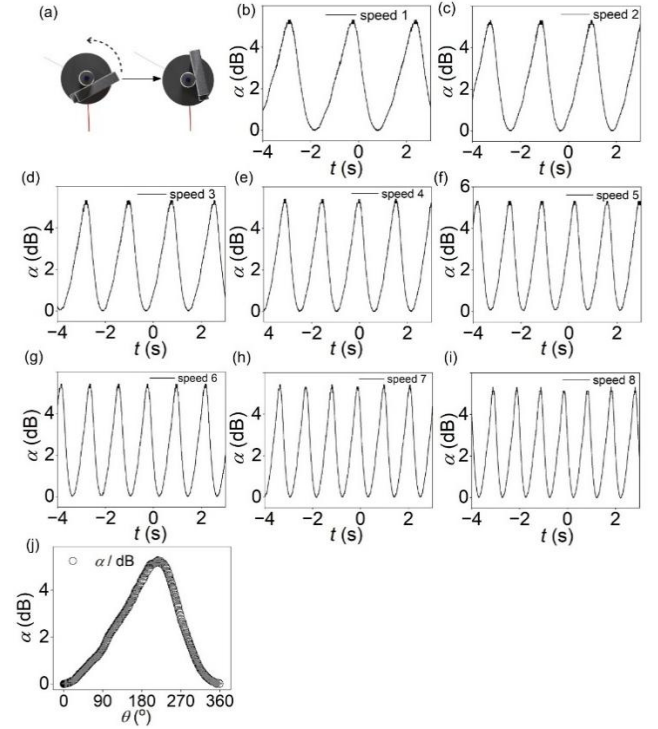

**Fig. 6.** Measured attenuation, $a$, from Port 3 (optical fibre length $L = 3$ m between Port 3 and power meter) during rotation of the magnet at relative displacement positions. Rotation is illustrated in Fig. 3(a) around the circulator set at different rotation velocities of $v = \Delta\theta/t = 135°/s$ to $v = 370$ °/s. A full 360° rotation returns attenuation back to $\alpha = 0$ dB; (b-i) shows the encoder response at different velocities; (j) the encoder system response calibrated against rotation angle, $q$.

After characterisation, the encoder is integrated into a 3D printed robot arm rotary joint, shown in Fig. 7 (a), to assess its potential in industrial environments, for robot body gesture sensing with rotation angle detection. The rotation angle of the robot arm is detected with the magneto-optical rotation encoder. For comparison, it is calibrated against a simple direct measurement of angle using a protractor. Fig. 7 (b) shows good agreement between the encoder and protractor measurements by a commercial protractor within $\Delta\theta \sim 0.5°$ and with a relatively small encoder standard deviation ($\sigma = 0.5°$). These discrepancies could be further minimised through on-site calibration prior to deployment in practical applications.

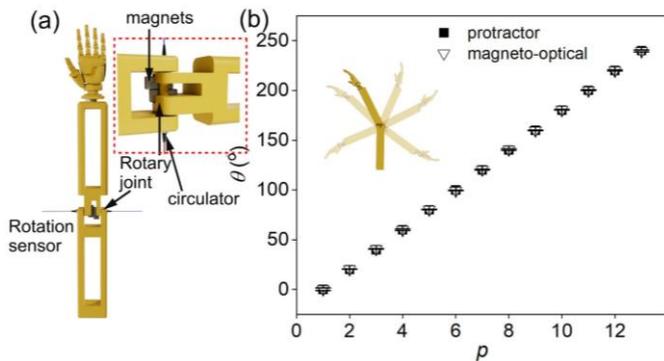

**Fig. 7.** (a) The circulator-based rotation encoder is integrated into a robot arm for monitoring and potentially controlling robot movement is inserted into a 3D printed robot arm rotary joint; (b) the derived rotation angle, $q$, measured by the encoder (black square) where errors $\sigma \sim 0.2°$ to $1.0°$ with an average of $\sigma \sim 0.5°$, compared with the angle read from a protractor (open triangle), where error is $\sigma \sim \pm 1.0°$, as a function of robot rotation increment, $p$.

## III. CONCLUSION

In conclusion, a magneto-optical encoder that combines both the advantages of optical fibre transmission with the modified complex functionality of a telecommunications optical circulator has been demonstrated. A two-dimensional robot arm was monitored over a full angular dynamic rotation range of $\theta = 0$ to $360°$ with a high resolution of $\Delta\theta = 0.3°$. After optimising the relative positions of additional magnets to help resist any further external interferences through a differential design, supported by simulation, the encoder achieves stable performance across a range of rotational speeds from $v = 135$ °/s to $v = 370$ °/s. This was demonstrated for a simple 3D printed robot arm, but such encoders have wider applications beyond industrial and medical robots, such as electric vehicle wheels, steering, motor rotors, aircraft engines, and more. Further work to improve accuracy and reduce noise in custom application environments would use additional refinements of differential sensing with two or more fibre encoders on each side of the joint. The work can be extended to more arbitrary degrees of freedom by adding angular considerations to eq. (1) and combining two encoders analysed by machine learning or AI routines that would decode such movement in a more complex freeform robot. Such technology also has applications in gyroscopes and other orientation technologies.

## REFERENCES

[1] S. T. Woo, Y. B. Park, J. H. Lee, C. S. Han, S. Na, and J. Y. Kim, 'Angle Sensor Module for Vehicle Steering Device Based on Multi-Track Impulse Ring', *Sensors*, vol. 19, no. 3, Art. no. 3, Jan. 2019, doi: 10.3390/s19030526.

[2] W. Kim, C. M. Kang, Y.-S. Son, and C. C. Chung, 'Nonlinear Steering Wheel Angle Control Using Self-Aligning Torque with Torque and Angle Sensors for Electrical Power Steering of Lateral Control System in Autonomous Vehicles', *Sensors*, vol. 18, no. 12, p. 4384, Dec. 2018, doi: 10.3390/s18124384.

[3] K. Shang, Y. Zhang, M. Galea, V. Brusic, and S. Korposh, 'Fibre optic sensors for the monitoring of rotating electric machines: a review', *Opt Quant Electron*, vol. 53, no. 2, p. 75, Jan. 2021, doi: 10.1007/s11082-020-02712-y.

[4] Z. Wang et al., 'A Self-Powered Angle Sensor at Nanoradian-Resolution for Robotic Arms and Personalized Medicare', *Advanced Materials*, vol. 32, no. 32, p. 2001466, 2020, doi: 10.1002/adma.202001466.

[5] J. Mata-Contreras, C. Herrojo, and F. Martín, 'Application of Split Ring Resonator (SRR) Loaded Transmission Lines to the Design of Angular Displacement and Velocity Sensors for Space Applications', *IEEE Transactions on Microwave Theory and Techniques*, vol. 65, no. 11, pp. 4450–4460, Nov. 2017, doi: 10.1109/TMTT.2017.2693981.

[6] M. El-Gohary and J. McNames, 'Human Joint Angle Estimation with Inertial Sensors and Validation with A Robot Arm', *IEEE Transactions on Biomedical Engineering*, vol. 62, no. 7, pp. 1759–1767, Jul. 2015, doi: 10.1109/TBME.2015.2403368.

[7] B. Siciliano and O. Khatib, 'Robotics and the Handbook', in *Springer Handbook of Robotics*, B. Siciliano and O. Khatib, Eds., Cham: Springer International Publishing, 2016, pp. 1–6. doi: 10.1007/978-3-319-32552-1_1.

[8] S. Wang, R. Ma, F. Cao, L. Luo, and X. Li, 'A Review: High-Precision Angle Measurement Technologies', *Sensors*, vol. 24, no. 6, Art. no. 6, Jan. 2024, doi: 10.3390/s24061755.

[9] M. Gasperi and P. "Philo" Hurbain, 'Potentiometer Sensors', in *Extreme NXT: Extending the LEGO MINDSTORMS NXT to the Next Level*, M. Gasperi and P. "Philo" Hurbain, Eds., Berkeley, CA: Apress, 2009, pp. 103–118. doi: 10.1007/978-1-4302-2454-9_6.

[10] G. Brasseur, 'Modeling of the front end of a new capacitive finger-type angular-position sensor', *IEEE Transactions on Instrumentation and Measurement*, vol. 50, no. 1, pp. 111–116, Feb. 2001, doi: 10.1109/19.903887.

[11] T. Sauter and N. Kero, 'System analysis of a fully-integrated capacitive angular sensor', *IEEE Transactions on Instrumentation and Measurement*, vol. 51, no. 6, pp. 1328–1333, Dec. 2002, doi: 10.1109/TIM.2002.808040.

[12] M. Gasulla, X. Li, G. C. M. Meijer, L. van der Ham, and J. W. Spronck, 'A contactless capacitive angular-position sensor', *IEEE Sensors Journal*, vol. 3, no. 5, pp. 607–614, Oct. 2003, doi: 10.1109/JSEN.2003.817182.

[13] D. Zheng, S. Zhang, S. Wang, C. Hu, and X. Zhao, 'A Capacitive Rotary Encoder Based on Quadrature Modulation and Demodulation', *IEEE Transactions on Instrumentation and Measurement*, vol. 64, no. 1, pp. 143–153, Jan. 2015, doi: 10.1109/TIM.2014.2328456.

[14] R. Pallás-Areny and J. G. Webster, *Sensors and Signal Conditioning*. John Wiley & Sons, 2012.

[15] P. Luo, Q. Tang, H. Jing, and X. Chen, 'Design and Development of a Self-Calibration- Based Inductive Absolute Angular Position Sensor', *IEEE Sensors Journal*, vol. 19, no. 14, pp. 5446–5453, Jul. 2019, doi: 10.1109/JSEN.2019.2908927.

[16] F. Kimura, M. Gondo, A. Yamamoto, and T. Higuchi, 'Resolver compatible capacitive rotary position sensor', in *2009 35th Annual Conference of IEEE Industrial Electronics*, Nov. 2009, pp. 1923–1928. doi: 10.1109/IECON.2009.5414847.

[17] V. Ferrari, A. Ghisla, D. Marioli, and A. Taroni, 'Capacitive angular-position sensor with electrically floating conductive rotor and measurement redundancy', *IEEE Transactions on Instrumentation and Measurement*, vol. 55, no. 2, pp. 514–520, Apr. 2006, doi: 10.1109/TIM.2006.864243.

[18] W. Lai, L. Cao, P. T. Phan, I.-W. Wu, S. C. Tjin, and S. Jay Phee, 'Joint Rotation Angle Sensing of Flexible Endoscopic Surgical Robots', in *2020 IEEE International Conference on Robotics and Automation (ICRA)*, May 2020, pp. 4789–4795. doi: 10.1109/ICRA40945.2020.9196549.

[19] Z. Li et al., 'Design and Application of Multidimensional Force/Torque Sensors in Surgical Robots: A Review', *IEEE Sensors Journal*, vol. 23, no. 12, pp. 12441–12454, Jun. 2023, doi: 10.1109/JSEN.2023.3270229.

[20] J. Zhao, W. Ou, N. Cai, Z. Wu, and H. Wang, 'Measurement Error Analysis and Compensation for Optical Encoders: A Review', *IEEE Transactions on Instrumentation and Measurement*, vol. 73, pp. 1–30, 2024, doi: 10.1109/TIM.2024.3417589.

[21] Z. Zhang, Y. Dong, F. Ni, M. Jin, and H. Liu, 'A Method for Measurement of Absolute Angular Position and Application in a

Novel Electromagnetic Encoder System', *Journal of Sensors*, vol. 2015, no. 1, p. 503852, 2015, doi: 10.1155/2015/503852.

[22] J. R. R. Mayer, 'High-resolution of rotary encoder analog quadrature signals', *IEEE Transactions on Instrumentation and Measurement*, vol. 43, no. 3, pp. 494–498, Jun. 1994, doi: 10.1109/19.293478.

[23] Y. Matsuzoe, N. Tsuji, T. Nakayama, K. Fujita, and T. Yoshizawa, 'High-performance absolute rotary encoder using multitrack and M-code', *OE*, vol. 42, no. 1, pp. 124–131, Jan. 2003, doi: 10.1117/1.1523943.

[24] P. Zu *et al.*, 'A Temperature-Insensitive Twist Sensor by Using Low-Birefringence Photonic-Crystal-Fiber-Based Sagnac Interferometer', *IEEE Photonics Technology Letters*, vol. 23, no. 13, pp. 920–922, Jul. 2011, doi: 10.1109/LPT.2011.2143400.

[25] D. K. Kim, J. Kim, S.-L. Lee, S. Choi, M. S. Kim, and Y. W. Lee, 'Twist-Direction-Discriminable Torsion Sensor Using Long-Period Fiber Grating Inscribed on Polarization-Maintaining Photonic Crystal Fiber', *IEEE Sensors Journal*, vol. 20, no. 6, pp. 2953–2961, Mar. 2020, doi: 10.1109/JSEN.2019.2954817.

[26] H. Y. Fu, S. K. Khijwania, H. Y. Tam, P. K. A. Wai, and C. Lu, 'Polarization-maintaining photonic-crystal-fiber-based all-optical polarimetric torsion sensor', *Appl. Opt., AO*, vol. 49, no. 31, pp. 5954–5958, Nov. 2010, doi: 10.1364/AO.49.005954.

[27] D. E. Ceballos-Herrera, I. Torres-Gómez, A. Martínez-Ríos, L. García, and J. J. Sánchez-Mondragón, 'Torsion Sensing Characteristics of Mechanically Induced Long-Period Holey Fiber Gratings', *IEEE Sensors Journal*, vol. 10, no. 7, pp. 1200–1205, Jul. 2010, doi: 10.1109/JSEN.2010.2042951.

[28] Y. Zhao, J. Shen, Q. Liu, and C. Zhu, 'Optical fiber sensor based on helical Fibers: A review', *Measurement*, vol. 188, p. 110400, Jan. 2022, doi: 10.1016/j.measurement.2021.110400.

[29] R. Gao, Y. Jiang, and L. Jiang, 'Multi-phase-shifted helical long period fiber grating based temperature-insensitive optical twist sensor', *Opt. Express, OE*, vol. 22, no. 13, pp. 15697–15709, Jun. 2014, doi: 10.1364/OE.22.015697.

[30] Z. Song, Y. Li, and J. Hu, 'Directional Torsion Sensor Based on a Two-Core Fiber with a Helical Structure', *Sensors*, vol. 23, no. 6, Art. no. 6, Jan. 2023, doi: 10.3390/s23062874.

[31] C. T. Lane, 'Kovar Glass Seals at Liquid Helium Temperatures', *Rev. Sci. Instrum.*, vol. 20, no. 2, pp. 140–140, Feb. 1949, doi: 10.1063/1.1741472.

[32] P. W. Droll and E. J. Iufer, 'Magnetic properties of selected spacecraft materials', 1967.

[33] Y. Guo, J. Canning, Z. Chaczko, and G.-D. Peng, 'Compact, remote optical waveguide magnetic field sensing using double-pass Faraday rotation-induced optical attenuation', *Appl. Opt., AO*, vol. 63, no. 14, pp. D35–D40, May 2024, doi: 10.1364/AO.513826.

[34] Y. Guo, J. Canning, Z. Chaczko, and G.-D. Peng, 'Magneto-optic fiber-coupled tuneable optical attenuation', *Opt. Lett., OL*, vol. 50, no. 2, pp. 277–280, Jan. 2025, doi: 10.1364/OL.543879.

[35] C. R. Harris *et al.*, 'Array programming with NumPy', *Nature*, vol. 585, no. 7825, pp. 357–362, Sep. 2020, doi: 10.1038/s41586-020-2649-2.

[36] T. E. Oliphant, 'Python for Scientific Computing', *Computing in Science & Engineering*, vol. 9, no. 3, pp. 10–20, May 2007, doi: 10.1109/MCSE.2007.58.

[37] J. D. Hunter, 'Matplotlib: A 2D Graphics Environment', *Computing in Science & Engineering*, vol. 9, no. 03, pp. 90–95, May 2007, doi: 10.1109/MCSE.2007.55.

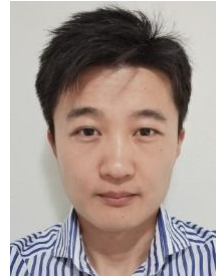

**Yunlong Guo** received the Ph.D. degree in Photonics from the University of New South Wales, Sydney, Australia. His doctoral research concentrated on the development of magneto-optical fiber sensors utilizing the Faraday effect, with applications in angular displacement measurement and magnetic field detection. His research interests further extend to optical sensing and electro-optical transducer technologies, with an emphasis on designing robust optical devices for use in harsh environments and potential applications in environmental monitoring and biomedical diagnostics.

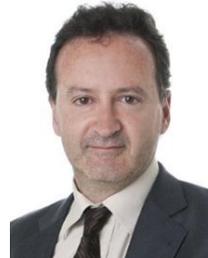

**John Canning** is a Fellow of SPIE, Optica, and the Governance Institute of Australia. He was an Australian Research Council Fellow, Future Fellow and Professorial Fellow, an Otto Monsted Professor at DTU Denmark (2004), a Villum Kann Rasmussen Professor at Aarhus University (2007), and a Science Without Borders Professor in Brazil (2009 - 2016), and an Honorary Professor at Montpelier Polytechnic, France (2019 - 2022). He has co-founded several companies and advised others, has over 850 journal and conference publications, over thirty patents, and was awarded the Optica David Richardson Medal in 2017 and a Changjiang Scholar distinguished professorship at the University of South China in China in 2025. He works as an expert consultant and has advised on policy and research culture as well as international collaborations.

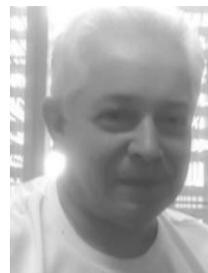

**Zenon Chaczko** received the bachelor's degree (Hons.) in computer science and the Ph.D. degree in computer and software engineering. He is currently a Senior Lecturer and the Director of the ICT Programs with the Faculty of Engineering and Information Technology, University of Technology, Sydney (UTS). He is also a Core Member of GBDTC–Global Big Data Technologies Center. Before UTS, he worked full-time for over 20 years in the software and system engineering industry. He held various visiting positions: visiting professorships at the University of Arizona, Tucson, USA, IPN, Mexico City, Mexico, the University of Las Palmas, Canary Islands, Spain, the Wroclaw University of Technology, Poland, and the University of Applied Science and Engineering, Hagenberg/Linz, Austria. He was a keynote speaker at multiple conferences. He has been actively conducting research in his fields and has published over 250 research articles. His research interests include artificial intelligence, software systems engineering, cloud and ambient computing (wireless sensor/actuators networks, the Internet of Things (IoT), and body area networks), augmented reality, and complex software systems (laparoscopic simulation, early detection of environmental/medical anomalies, and morphotronics).

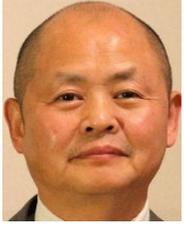 **Gang-Ding Peng** received the B.Sc. degree in physics from Fudan University, Shanghai, China, in 1982, and the M.Sc. and Ph.D. degrees in electronic engineering from Shanghai Jiao Tong University, Shanghai, in 1984 and 1987, respectively. He has been working with the University of New South Wales, Sydney, NSW, Australia, since 1991, where he was a Queen Elizabeth II Fellow from 1992 to 1996 and is currently a professor. His research interests include silica and polymer optical fibers, optical fiber and waveguide devices, optical fiber sensors, and nonlinear optics. Dr. Peng is a Fellow and a Life Member of the Optical Society of America (OSA) and Society of Photo-Optical Instrumentation Engineers (SPIE).